\documentclass[runningheads]{llncs}

\usepackage{amsfonts}
\usepackage{amsmath}
\usepackage{graphicx}
\usepackage{hyperref}
\usepackage{marvosym}
\usepackage{multirow}
\usepackage{subfig}

\begin{document}

\pagestyle{empty}
\mainmatter

%--------------------
\title{Cross-lingual Entity Alignment via\\ Joint Attribute-Preserving Embedding}
\titlerunning{Cross-lingual Entity Alignment via JAPE}

\author{Zequn Sun\inst{1} \and Wei Hu\inst{1}\textsuperscript{(\Letter)} \and Chengkai Li\inst{2}}
\authorrunning{Z. Sun et al.}

\institute{State Key Laboratory for Novel Software Technology,\\ Nanjing University, China\\
  \email{zqsun.nju@gmail.com, whu@nju.edu.cn}\\
 \and Department of Computer Science and Engineering,\\ University of Texas at Arlington, USA\\
  \email{cli@uta.edu}}
  
\maketitle

%--------------------

\begin{abstract}
Entity alignment is the task of finding entities in two knowledge bases (KBs) that represent the same real-world object. When facing KBs in different natural languages, conventional cross-lingual entity alignment methods rely on machine translation to eliminate the language barriers. These approaches often suffer from the uneven quality of translations between languages. While recent embedding-based techniques encode entities and relationships in KBs and do not need machine translation for cross-lingual entity alignment, a significant number of attributes remain largely unexplored. In this paper, we propose a joint attribute-preserving embedding model for cross-lingual entity alignment. It jointly embeds the structures of two KBs into a unified vector space and further refines it by leveraging attribute correlations in the KBs. Our experimental results on real-world datasets show that this approach significantly outperforms the state-of-the-art embedding approaches for cross-lingual entity alignment and could be complemented with methods based on machine translation.
\keywords{cross-lingual entity alignment, knowledge base embedding, joint attribute-preserving embedding}
\end{abstract}

%--------------------

\section{Introduction}
\label{sect:intro}

In the past few years, knowledge bases (KBs) have been successfully used in lots of AI-related areas such as Semantic Web, question answering and Web mining. Various KBs cover a broad range of domains and store rich, structured real-world facts.  
In a KB, each fact is stated in a triple of the form $(entity,property,value)$, in which $value$ can be either a literal or an entity. The sets of entities, properties, literals and triples are denoted by $E,P,L$ and $T$, respectively.  Blank nodes are ignored for simplicity. There are two types of properties---relationships ($R$) and attributes ($A$)---and correspondingly two types of triples, namely relationship triples and attribute triples. A relationship triple $tr\in E\times R\times E$ describes the relationship between two entities, e.g. $(Texas,hasCapital,Austin)$, while an attribute triple $tr\in E\times A\times L$ gives a literal attribute value to an entity, e.g. $(Texas,areaTotal,``696241.0")$.

As widely noted, KBs often suffer from two problems: (i) Low coverage. Different KBs are constructed by different parties using different data sources. They contain complementary facts, which makes it imperative to integrate multiple KBs. (ii) Multi-linguality gap. To support multi-lingual applications, a growing number of multi-lingual KBs and language-specific KBs have been built. This makes it both necessary and beneficial to integrate cross-lingual KBs. 

Entity alignment is the task of finding entities in two KBs that refer to the same real-world object. It plays a vital role in automatically integrating multiple KBs. This paper focuses on cross-lingual entity alignment. It can help construct a coherent KB and deal with different expressions of knowledge across diverse natural languages. Conventional cross-lingual entity alignment methods rely on machine translation, of which the accuracy is still far from perfect. Spohr {\it et al.} \cite{CrosslingualOM} argued that the quality of alignment in cross-lingual scenarios heavily depends on the quality of translations between multiple languages.

Following the popular translation-based embedding models~\cite{TransE,TransR,TransH}, a few studies leveraged KB embeddings for entity alignment and achieved promising results~\cite{MTransE,JE}. Embedding techniques learn low-dimensional vector representations (i.e., embeddings) of entities and encode various semantics (e.g. types) into them. Focusing on KB structures, the embedding-based methods provide an alternative for cross-lingual entity alignment without considering their natural language labels. 

There remain several challenges in applying embedding methods to cross-lingual entity alignment. First, to the best of our knowledge, most existing KB embedding models learn embeddings based solely on relationship triples. However, we observe that attribute triples account for a significant portion of KBs. For example, we count triples of infobox facts from English DBpedia (2016-04),\footnote{\url{http://wiki.dbpedia.org/downloads-2016-04}} and find 58,181,947 attribute triples, which are three times as many as relationship triples (the number is 18,598,409). Facing the task of entity alignment, attribute triples can provide additional information to embed entities, but how to incorporate them into cross-lingual embedding models remains largely unexplored. Second, thanks to the Linking Open Data initiative, there exist some aligned entities and properties between KBs, which can serve as bridge between them. However, as discovered in \cite{MTransE}, the existing alignment between cross-lingual KBs usually accounts for a small proportion. So how to make the best use of it is crucial for embedding cross-lingual KBs. 

To deal with the above challenges, we introduce a joint attribute-preserving embedding model for cross-lingual entity alignment. It employs two modules, namely structure embedding (SE) and attribute embedding (AE), to learn embeddings based on two facets of knowledge (relationship triples and attribute triples) in two KBs, respectively. SE focuses on modeling relationship structures of two KBs and leverages existing alignment given beforehand as bridge to overlap their structures. AE captures the correlations of attributes (i.e. whether these attributes are commonly used together to describe an entity) and clusters entities based on attribute correlations. Finally, it combines SE and AE to jointly embed all the entities in the two KBs into a unified vector space $\mathbb{R}^d$, where $d$ denotes the dimension of the vectors. The aim of our approach is to find latent cross-lingual target entities (i.e. truly-aligned entities that we want to discover) for a source entity by searching its nearest neighbors in $\mathbb{R}^d$. We expect the embeddings of latent aligned cross-lingual entities to be close to each other. 

In summary, the main contributions of this paper are as follows:
\begin{itemize}
\item We propose an embedding-based approach to cross-lingual entity alignment, which does not depend on machine translation between cross-lingual KBs.

\item We jointly embed the relationship triples of two KBs with structure embedding and further refine the embeddings by leveraging attribute triples of KBs with attribute embedding. To the best of our knowledge, there is no prior work learning embeddings of cross-lingual KBs while preserving their attribute information.

\item We evaluated our approach on real-world cross-lingual datasets from DBpedia. The experimental results show that our approach largely outperformed two state-of-the-art embedding-based methods for cross-lingual entity alignment. Moreover, it could be complemented with conventional methods based on machine translation.
\end{itemize}

The rest of this paper is organized as follows. %We begin with the preliminaries in Section~\ref{sect:pre}, where we formalize the problem and describe notations. 
We discuss the related work on KB embedding and cross-lingual KB alignment in Section \ref{sect:work}. We describe our approach in detail in Section~\ref{sect:method}, and report experimental results in Section~\ref{sect:eval}. Finally, we conclude this paper with future work in Section~\ref{sect:concl}.

%\section{Preliminaries}
%\label{sect:pre}

%A knowledge base, denoted by $KB$, consists of a set of $(entity, property, value)$ triples, where $value$ can be a literal or an entity. The sets of entities, properties, literals and triples in $KB$ are denoted by $E,P,L$ and $T$, respectively. We refer to a triple $tr\in E\times P\times E$ as a relationship triple and denote the set of relationship properties by $R$. An attribute triple is in $E\times P\times L$, and the set of attribute properties is denoted by $A$. Blank nodes are ignored for simplicity.

%Cross-lingual entity alignment is the task of finding entities in two KBs that refer to the same real-world object, where the two KBs are labeled in different natural languages. In this paper, we specifically use the embedding techniques to learn entity representations in vector space $\mathbb{R}^d$, and find latent cross-lingual target entities for a source entity by searching its nearest neighbors in $\mathbb{R}^d$, as we expect the latent aligned entities to be embedded closely.

%Lowercase boldfaced letters denote vector representations of the corresponding terms, e.g. $(\mathbf{h},\mathbf{r},\mathbf{t})$ denotes the vector representation of triple $(h,r,t)$. Capital boldfaced letters denote representation matrices. We use superscripts to distinguish notations of different KBs. For example, $\mathbf{E}^{(1)}$ denotes the representation matrix for entities in $KB_1$ with each row an entity vector $\mathbf{e}^{(1)}$.

\section{Related Work}
\label{sect:work}

We divide the related work into two subfields: KB embedding and cross-lingual KB alignment. We discuss them in the rest of this section.

\subsection{KB Embedding}

In recent years, significant efforts have been made towards learning embeddings of KBs. TransE \cite{TransE}, the pioneer of translation-based methods, interprets a relationship vector as the translation from the head entity vector to its tail entity vector. In other words, if a relationship triple $(h,r,t)$ holds, $\mathbf{h}+\mathbf{r}\approx\mathbf{t}$ is expected. TransE has shown its great capability of modeling 1-to-1 relations and achieved promising results for KB completion. To further improve TransE, later work including TransH \cite{TransH} and TransR \cite{TransR} was proposed. Additionally, there exist a few non-translation-based approaches to KB embedding \cite{SE,RESCAL,NTN}.

Besides, several studies take advantage of knowledge in KBs to improve embeddings. Krompa\ss\ {\it et al.} \cite{TypeConstrain} added type constraints to KB embedding models and enhanced their performance on link prediction. KR-EAR \cite{AttributeEmbedding} embeds attributes additionally by modeling attribute correlations and obtains good results on predicting entities, relationships and attributes. But it only learns attribute embeddings in a single KB, which hinders its application to cross-lingual cases. Besides, KR-EAR focuses on the attributes whose values are from a small set of entries, e.g. values of ``gender" are \{Female, Male\}. It may fail to model attributes whose values are very sparse and heterogeneous, e.g. ``name", ``label" and ``coordinate". RDF2Vec \cite{RDF2Vec} uses local information of KB structures to generate sequences of entities and employs language modeling approaches to learn entity embeddings for machine learning tasks. For cross-lingual tasks, \cite{KBCompletion} extends NTNKBC \cite{NTNKBC} for cross-lingual KB completion. \cite{KBTranslation} uses a neural network approach that translates English KBs into Chinese to expand Chinese KBs. 

\subsection{Cross-lingual KB Alignment}

Existing work on cross-lingual KB alignment generally falls into two categories: cross-lingual ontology matching and cross-lingual entity alignment. For cross-lingual ontology matching, Fu {\it et al.} \cite{Survey,MultilingualSW} presented a generic framework, which utilizes machine translation tools to translate labels to the same language and uses monolingual ontology matching methods to find mappings. Spohr {\it et al.} \cite{CrosslingualOM} leveraged translation-based label similarities and ontology structures as features for learning cross-lingual mapping functions by machine learning techniques (e.g. SVM). In all these works, machine translation is an integral component.

For cross-lingual entity alignment, MTransE \cite{MTransE} incorporates TransE to encode KB structures into language-specific vector spaces and designs five alignment models to learn translation between KBs in different languages with seed alignment. JE \cite{JE} utilizes TransE to embed different KBs into a unified space with the aim that each seed alignment has similar embeddings, which is extensible to the cross-lingual scenario. Wang {\it et al.} \cite{Wiki} proposed a graph model, which only leverages language-independent features (e.g. out-/inlinks) to find cross-lingual links between Wiki knowledge bases. Gentile {\it et al.} \cite{TableEmbedding} exploited embedding-based methods for aligning entities in Web tables. Different from them, our approach jointly embeds two KBs together and leverages attribute embedding for improvement.

\section{Cross-lingual Entity Alignment via KB Embedding}
\label{sect:method}

In this section, we first introduce notations and the general framework of our joint attribute-preserving embedding model. Then, we elaborate on the technical details of the model and discuss several key design issues.

We use lower-case bold-face letters to denote the vector representations of the corresponding terms, e.g., $(\mathbf{h},\mathbf{r},\mathbf{t})$ denotes the vector representation of triple $(h,r,t)$. We use capital bold-face letters to denote matrices, and we use superscripts to denote different KBs. For example, $\mathbf{E}^{(1)}$ denotes the representation matrix for entities in $KB_1$ in which each row is an entity vector $\mathbf{e}^{(1)}$. 

\subsection{Overview}

\begin{figure}[!tbp]
\centering
\includegraphics[width=\columnwidth]{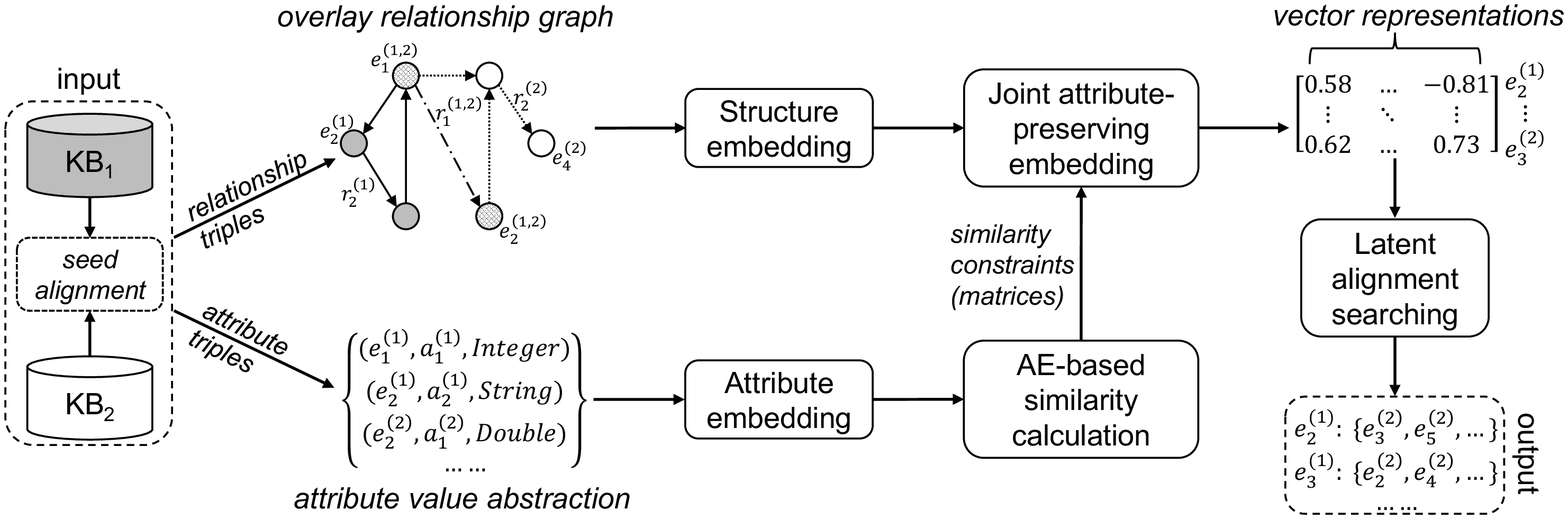}
\caption{Framework of the joint attribute-preserving embedding model}
\label{fig:framework}
\end{figure}

The framework of our joint attribute-preserving embedding model is depicted in Fig.~\ref{fig:framework}. Given two KBs, denoted by $KB_1$ and $KB_2$, in different natural languages and some pre-aligned entity or property pairs (called seed alignment, denoted by superscript $^{(1,2)}$), our model learns the vector representations of $KB_1$ and $KB_2$ and expects the latent aligned entities to be embedded closely. 

Following TransE \cite{TransE}, we interpret a relationship as the translation from~the head entity to the tail entity, to characterize the structure information of KBs. We let each pair in the seed alignment share the same representation to serve as bridge between $KB_1$ and $KB_2$ to build an overlay relationship graph, and learn representations of all the entities jointly under a unified vector space via structure embedding (SE). The intuition is that two alignable KBs are likely to have a number of aligned triples, e.g. $(Washington,capitalOf,America)$ in English and its correspondence $(Washington,capitaleDes,\acute{E}tats\text{-}Unis)$ in French. Based on this, SE aims at learning approximate representations for the latent aligned triples between the two KBs.

However, SE only constrains that the learned representations must be compatible within each relationship triple, which causes the disorganized distribution of some entities due to the sparsity of their relationship triples. To alleviate this incoherent distribution, we leverage attribute triples for helping embed entities based on the observation that the latent aligned entities usually have a high degree of similarity in attribute values. Technically, we overlook specific attribute values by reason of their complexity, heterogeneity and cross-linguality. Instead, we abstract attribute values to their range types, e.g. $(Tom,age,``12")$ to $(Tom,age,Integer)$, where $Integer$ is the abstract range type of value ``12". Then, we carry out attribute embedding (AE) on abstract attribute triples to capture the correlations of cross-lingual and mono-lingual attributes, and calculate the similarities of entities based on them. Finally, the attribute similarity constraints are combined with SE to refine representations by clustering entities with high attribute correlations. In this way, our joint model preserves both relationship and attribute information of the two KBs.

With entities represented as vectors in a unified embedding space, the alignment of latent cross-lingual target entities for a source entity can be conducted by searching the nearest cross-lingual neighbors in this space.

\subsection{Structure Embedding}
\label{subsect:se}

The aim of SE is to model the geometric structures of two KBs and learn approximate representations for latent aligned triples. Formally, given a relationship triple $tr=(h,r,t)$, we expect $\mathbf{h}+\mathbf{r}=\mathbf{t}$. To measure the plausibility of $tr$, we define the score function $f(tr)={\|\mathbf{h}+\mathbf{r}-\mathbf{t}\|}_2^2$. We prefer a lower value of $f(tr)$ and want to minimize it for each relationship triple.

\begin{figure}[!tbp]
\centering
\includegraphics[width=1\columnwidth]{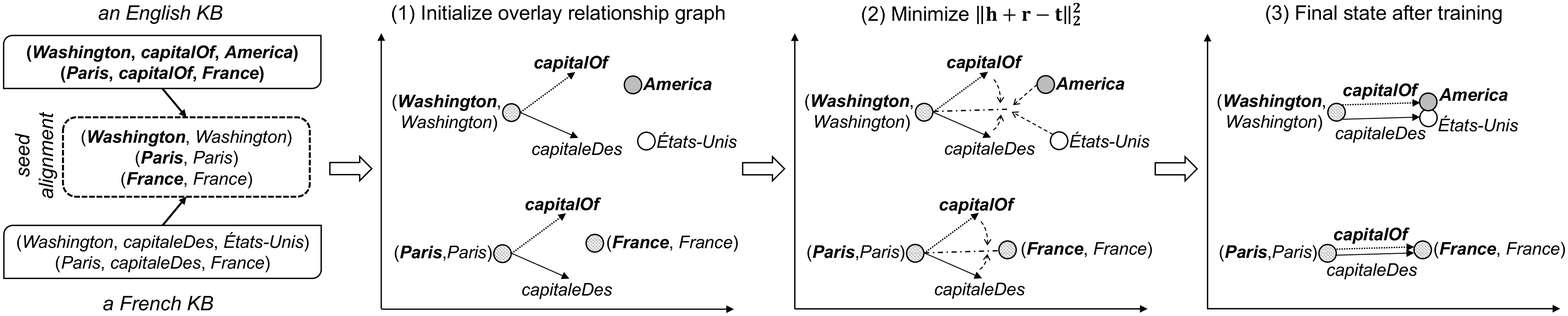}
\caption{An example of structure embedding}
\label{fig:example}
\end{figure}

Fig~\ref{fig:example} gives an example about how SE models the geometric structures of two KBs with seed alignment. In Phase (1), we initialize all the vectors randomly and let each pair in seed alignment overlap to build the overlay relationship graph. In order to show the triples intuitively in the figure, we regard an entity as a point in the vector space and move relationship vectors to start from their head entities. Note that, currently, entities and relationships distribute randomly. In Phase (2), we minimize scores of triples and let vector representations compatible within each relationship triple. For example, the relationship $capitalOf$ would tend to be close to $capitaleDes$ because they share the same head entity and tail entity. In the meantime, the entity $America$ and its correspondence $\acute{E}tats\text{-}Unis$ would move closely to each other due to their common head entity and approximate relationships. Therefore, SE is a dynamic spreading process. The ideal state after training is shown as Phase (3). We can see that the latent aligned entities $America$ and $\acute{E}tats\text{-}Unis$ lie together.

Furthermore, we detect that negative triples (a.k.a. corrupted triples), which have been widely used in translation-based embedding models \cite{TransE,TransR,TransH}, are also valuable to SE. Considering that another English entity $China$ and its latent aligned French one $Chine$ happen to lie closely to $America$, SE may take the $Chine$ as a candidate for $America$ by mistake due to their short distance. Negative triples would help reduce the occurrence of this coincidence. If we generate a negative triple $tr' = (Washington,capitalOf,China)$ and learn a high score for $tr'$, $China$ would keep a distance away from $America$. As we enforce the length of any embedding vector to $1$,  the score function $ f $ has a constant maximum. Thus, we would like to minimize $ -f(tr') $ to learn a high score for $ tr' $.

In summary, we prefer lower scores for existing triples (positives) and higher scores for negatives, which leads to minimize the following objective function:
\begin{equation}
\label{eq:se}
\mathcal{O}_{SE}=\sum_{tr\in T}\sum_{tr'\in T'_{tr}} \big(f(tr)-\alpha f(tr')\big),
\end{equation}
where $T$ denotes the set of all positive triples and $T'_{tr}$ denotes the associated negative triples for $tr$ generated by replacing either its head or tail by a random entity (but not both at the same time). $\alpha$ is a ratio hyper-parameter that weights positive and negative triples and its range is $[0,1]$. It is important to remember that each pair in the seed alignment share the same embedding during training, in order to bridge two KBs.

\subsection{Attribute Embedding and Entity Similarity Calculation}

\subsubsection{Attribute Embedding}

We call a set of attributes correlated if they are commonly used together to describe an entity. For example, attributes $longitude$, $latitude$ and $place\_name$ are correlated because they are widely used together to describe a place. Moreover, we want to assign a higher correlation to the pair of $longitude$ and $latitude$ because they have the same range type. We use seed entity pairs to establish correlations between cross-lingual attributes. Given an aligned entity pair $(e^{(1)},e^{(2)})$, we regard the attributes of $e^{(1)}$ as correlated ones for each attribute of $e^{(2)}$, and vice versa. We expect attributes with high correlations to be embedded closely.

To capture the correlations of attributes, AE borrows the idea from Skip-gram \cite{SkipGram}, a very popular model that learns word embeddings by predicting the context of a word given the word itself. Similarly, given an attribute, AE wants to predict its correlated attributes. In order to leverage the range type information, AE minimizes the following objective function:
\begin{equation} 
\label{eq:log}
\mathcal{O}_{AE}=-\sum_{(a,c)\in H}w_{a,c}\cdot\log p(c|a),
\end{equation}
where $H$ denotes the set of positive $(a,c)$ pairs, i.e., $c$ is actually a correlated attribute of $a$, and the term $ p(c|a)$ denotes the probability. To prevent all the vectors from having the same value, we adopt the negative sampling approach \cite{NegativeSampling} to efficiently parameterize Eq. (\ref{eq:log}), and $\log p(c|a)$ is  replaced with the term as follows:
\begin{equation} 
\label{eq:log_nce}
\log\sigma(\mathbf{a}\cdot\mathbf{c})+\sum_{(a,c')\in H_a'}\log\sigma(\mathbf{-a}\cdot\mathbf{c}'),
\end{equation}
where $\sigma(x)=\frac{1}{1+e^{-x}}$. $H_a'$ is the set of negative pairs for attribute $a$ generated according to a log-uniform base distribution, assuming that they are all incorrect.

We set $w_{a,c}=1$ if $a$ and $c$ have different range types, otherwise $w_{a,c}=2$ to increase their probability of tending to be similar. In this paper, we distinguish four kinds of abstract range types, i.e., $Integer,$ $Double, Datetime$ and $String$ (as default). Note that it is easy to extend to more types. 

%To assign weights for abstract range types, we empirically set $w_{a,c}=2$ if $a$ and $c$ have the same range type. It is also a alternative to borrow ideas from TF-IDF and let the weight of a type vary inversely as its frequency of occurrence.

\subsubsection{Entity Similarity Calculation}

Given attribute embeddings, we take the representation of an entity to be the normalized average of its attribute vectors, i.e., $\mathbf{e}={[\sum_{a\in A_e}\mathbf{a}]}_1$, where $A_e$ is the set of attributes of $e$ and $[\mathbf{.}]_1$ denotes the normalized vector. We have two matrices of vector representations for entities in two KBs, $\mathbf{E}_{AE}^{(1)}\in\mathbb{R}^{n_e^{(1)}\times d}$ for $KB_1$ and $\mathbf{E}_{AE}^{(2)}\in\mathbb{R}^{n_e^{(2)}\times d}$ for $KB_2$, where each row is an entity vector, and $n_e^{(1)},$ $n_e^{(2)}$ are the numbers of entities in $KB_1,KB_2$, respectively. 

We use the cosine distance to measure the similarities between entities. For two entities $e,e'$, we have $ {\rm sim}(e,e')={\cos} (\mathbf{e},\mathbf{e'})=\frac{\mathbf{e}\cdot \mathbf{e'}}{||\mathbf{e}||||\mathbf{e'}||}=\mathbf{e}\cdot\mathbf{e'}$, as the length of any embedding vector is enforced to 1. The cross-KB similarity matrix $\mathbf{S}^{(1,2)}\in\mathbb{R}^{n_e^{(1)}\times n_e^{(2)}}$ between $KB_1$ and $KB_2$, as well as the inner similarity matrices $\mathbf{S}^{(1)}\in\mathbb{R}^{n_e^{(1)}\times n_e^{(1)}}$ for $KB_1$ and $\mathbf{S}^{(2)}\in\mathbb{R}^{n_e^{(2)}\times n_e^{(2)}}$ for $KB_2$, are defined as follows:
\begin{equation} 
\label{eq:entsim}
\mathbf{S}^{(1,2)}=\mathbf{E}_{AE}^{(1)}{\mathbf{E}_{AE}^{(2)\top}},\quad 
\mathbf{S}^{(1)}=\mathbf{E}_{AE}^{(1)}{\mathbf{E}_{AE}^{(1)\top}},\quad 
\mathbf{S}^{(2)}=\mathbf{E}_{AE}^{(2)}{\mathbf{E}_{AE}^{(2)\top}}.
\end{equation}

A similarity matrix $\mathbf{S}$ holds the cosine similarities among entities and $\mathbf{S}_{i,j}$ is the similarity between the $i$-th entity in one KB and the $j$-th entity in the same or the other KB. We discard lower values of $\mathbf{S}$ because a low similarity of two entities indicates that they are likely to be different. So, we set the entry $\mathbf{S}_{i,j}=0$ if $\mathbf{S}_{i,j}<\tau$, where $\tau$ is a threshold and can be set based on the average similarity of seed entity pairs. In this paper, we fix $\tau=0.95$ for inner similarity matrices and $ 0.9 $ for cross-KB similarity matrix, to achieve high accuracy.

\subsection{Joint Attribute-Preserving Embedding}

We want similar entities across KBs to be clustered to refine their vector representations. Inspired by \cite{WordEmbedding}, we use the matrices of pairwise similarities between entities as supervised information and minimize the following objective function:
\begin{eqnarray} 
\label{eq:ae}
\mathcal{O}_{S} &=& {\|\mathbf{E}_{SE}^{(1)}-\mathbf{S}^{(1,2)}\mathbf{E}_{SE}^{(2)}\|}_F^2 \nonumber\\
&& +\,\beta({\|\mathbf{E}_{SE}^{(1)}-\mathbf{S}^{(1)}\mathbf{E}_{SE}^{(1)}\|}_F^2+{\|\mathbf{E}_{SE}^{(2)}-\mathbf{S}^{(2)}\mathbf{E}_{SE}^{(2)}\|}_F^2),
\end{eqnarray}
where $\beta$ is a hyper-parameter that balances similarities between KBs and their inner similarities. $\mathbf{E}_{SE} \in \mathbb{R}^{n_e \times d}$ denotes the matrix of entity vectors for one KB in SE with each row an entity vector. $\mathbf{S}^{(1,2)}\mathbf{E}_{SE}^{(2)}$ calculates latent vectors of entities in $KB_1$ by accumulating vectors of entities in $KB_2$ based on their similarities. By minimizing ${\|\mathbf{E}_{SE}^{(1)}-\mathbf{S}^{(1,2)}\mathbf{E}_{SE}^{(2)}\|}_F^2$, we expect similar entities across KBs to be embedded closely. The two inner similarity matrices work in the same way. 

To preserve both the structure and attribute information of two KBs, we jointly minimize the following objective function:
\begin{equation}
\label{eq:joint}
\mathcal{O}_{joint}=\mathcal{O}_{SE}+\delta\mathcal{O}_{S},
\end{equation}
where $\delta$ is a hyper-parameter weighting $\mathcal{O}_{S}$.

\subsection{Discussions}
\label{subsect:discuss}

We discuss and analyze our joint attribute-preserving embedding model in the following aspects:

%\subsubsection{Negative Triples}
%In contrast to MTransE \cite{MTransE} that does not use any negative triples, we argue that they are effective in distinguishing the relations between entities. Considering that there are two triples $(h,r_1,t_1)$ and $(h,r_2,t_2)$ with the same head entity $h$, if $t_1$ and $t_2$ happen to have similar vector representations, MTransE (using only positive triples for training) would accept this structure, which, however, has negative influence on the entity alignment task. We prefer the structure that can separate $t_1,t_2$, and negative triples would help reduce the occurrence of this coincidence. Our experimental results reported in Section \ref{subsect:result} also demonstrate the effectiveness of negative triples. 

\subsubsection{Objective Function for Structure Embedding}
SE is translation-based embedding model but its objective function (see Eq.~(\ref{eq:se})) does not follow the margin-based ranking loss function below, which is used by many previous KB embedding models \cite{TransE}:
\begin{equation} \label{eq:transe}
\mathcal{O}=\sum_{tr\in T}\sum_{tr'\in T'_{tr}} \max[\gamma+f(tr)-f(tr'),0].
\end{equation}

Eq.~(\ref{eq:transe}) aims at distinguishing positive and negative triples, and expects that their scores can be separated by a large margin. However, for the cross-lingual entity alignment task, in addition to the large margin between their scores, we also want to assign lower scores to positive triples and higher scores to negative triples. Therefore, we choose Eq.~(\ref{eq:se}) instead of Eq.~(\ref{eq:transe}).

In contrast, JE \cite{JE} uses the margin-based ranking loss from TransE \cite{TransE}, while MTransE \cite{MTransE} does not have this as it does not use negative triples. However, as explained in Section \ref{subsect:se}, we argue that negative triples are effective in distinguishing the relations between entities. Our experimental results reported in Section \ref{subsect:result} also demonstrate the effectiveness of negative triples. 

\subsubsection{Training}
We initialize parameters such as vectors of entities, relations and attributes randomly based on a truncated normal distribution, and then optimize Eqs.~(\ref{eq:log}) and (\ref{eq:joint}) with a gradient descent optimization algorithm called AdaGrad \cite{AdaGrad}. Instead of directly optimizing $\mathcal{O}_{joint}$, our training process involves two optimizers to minimize $\mathcal{O}_{SE}$ and $\delta\mathcal{O}_{S}$ independently. At each epoch, the two optimizers are executed alternately. When minimizing $\mathcal{O}_{SE}$, $ f(tr) $ and $ -\alpha f(tr') $ can also be optimized alternately.

The length of any embedding vector is enforced to 1 for the following reasons: (i) this constraint prevents the training process from trivially minimizing the objective function by increasing the embedding norms and shaping the embeddings, (ii) it limits the randomness of entity and relationship distribution in the training process, and (iii) it fixes the mismatch between the inner product in Eq. (\ref{eq:log_nce}) and the cosine similarity to measure embeddings \cite{Orthogonality}.

Our model is also scalable in training. The structure embedding belongs to the translation-based embedding models, which have already been proved to be capable of learning embeddings at large scale \cite{TransE}. We use sparse representations for matrices in Eq. (\ref{eq:ae}) for saving memory. Additionally, the memory cost to compute Eq.~(\ref{eq:entsim}) can be reduced using a divide-and-conquer strategy. 

\subsubsection{Parameter Complexity}
The number of parameters in our joint model is $d(n_e+n_r+n_a)$, where $n_e,n_r,n_a$ are the numbers of entities, relationships and attributes, respectively. $d$ is the dimension of the embeddings. Considering that $ n_r,n_a\ll n_e $ in practice and the seed alignment share vectors in training, the complexity of the model is roughly linear to the number of total entities. 

\subsubsection{Searching Latent Aligned Entities} 
Because the length of each vector always equals 1, the cosine distance between entities of the two KBs can be calculated as $\mathbf{D} = \mathbf{E}_{SE}^{(1)} {\mathbf{E}_{SE}^{(2)\top}}$. Thus, the nearest entities can be obtained by simply sorting each row of $ \mathbf{D} $ in descending order. For each source entity, we expect the rank of its truly-aligned target entity to be the first few.

%--------------------

\section{Evaluation}
\label{sect:eval}

In this section, we report our experiments and results on real-world cross-lingual datasets. We developed our approach, called JAPE, using TensorFlow\footnote{\url{https://www.tensorflow.org/}}---a very popular open-source software library for numerical computation. Our experiments were conducted on a personal workstation with an Intel Xeon E3 3.3 GHz CPU and 128 GB memory. The datasets, source code and experimental results are accessible at this website\footnote{\url{https://github.com/nju-websoft/JAPE}}.

\subsection{Datasets}

We selected DBpedia (2016-04) to build three cross-lingual datasets. DBpedia is a large-scale multi-lingual KB including inter-language links (ILLs) from entities of English version to those in other languages. In our experiments, we extracted 15 thousand ILLs with popular entities from English to Chinese, Japanese and French respectively, and considered them as our reference alignment (i.e., gold standards). Our strategy to extract datasets is that we randomly selected an ILL pair s.t. the involved entities have at least 4 relationship triples and then extracted relationship and attribute infobox triples for selected entities. The statistics of the three datasets are listed in Table~\ref{tab:stat}, which indicate that the number of involved entities in each language is much larger than 15 thousand, and attribute triples contribute to a significant portion of the datasets.

\begin{table}[!tbp]
 \centering
 \caption{Statistics of the datasets}
 \label{tab:stat}
 {\scriptsize \begin{tabular}{|l|l|rcccc|}
  \hline \multicolumn{2}{|c|}{Datasets} & Entities & Relationships & Attributes & Rel. triples & Attr. triples \\
  \hline \multirow{2}{*}{DBP15K$_\textrm{ZH-EN}$} & Chinese  & 66,469 & 2,830 & 8,113 & 153,929 & 379,684 \\
                                                  & English  & 98,125 & 2,317 & 7,173 & 237,674 & 567,755 \\                                               
  \hline \multirow{2}{*}{DBP15K$_\textrm{JA-EN}$} & Japanese & 65,744 & 2,043 & 5,882 & 164,373 & 354,619 \\
                                                  & English  & 95,680 & 2,096 & 6,066 & 233,319 & 497,230 \\                                           
  \hline \multirow{2}{*}{DBP15K$_\textrm{FR-EN}$} & French   & 66,858 & 1,379 & 4,547 & 192,191 & 528,665 \\
                                                  & English  & 105,889 & 2,209 & 6,422 & 278,590 & 576,543 \\                                                  
  %\hline \multirow{2}{*}{DBP15K$_\textrm{DE-EN}$} & German   & 48,670 & 1,006 & 2,836 & 127,523 & 363,150 \\
  %                                                & English  & 96,023 & 1,595 & 4,825 & 233,278 & 568,690 \\
  \hline 
\end{tabular}}
\end{table}

\subsection{Comparative Approaches}

As aforementioned, JE \cite{JE} and MTransE \cite{MTransE} are two representative embedding-based methods for entity alignment. In our experiments, we used our best effort to implement the two models as they do not release any source code or software currently. We conducted them on the above datasets as comparative approaches. Specifically, MTransE has five variants in its alignment model, where the fourth performs best according to the experiments of its authors. Thus, we chose this variant to represent MTransE. We followed the implementation details reported in \cite{MTransE,JE} and complemented other unreported details with careful consideration. For example, we added a strong orthogonality constraint for the linear transformation matrix in MTransE to ensure the invertibility, because we found it leads to better results. For JAPE, we tuned various parameter values and set $d=75,\alpha=0.1,\beta=0.05,\delta=0.05$ for the best performance. The learning rates of SE and AE were empirically set to 0.01 and 0.1, respectively.  

\subsection{Evaluation Metrics}

Following the conventions \cite{TransE,MTransE,JE}, we used $Hits@k$ and $Mean$ to assess the performance of the three approaches. $Hits@k$ measures the proportion of correctly aligned entities ranked in the top $k$, while $Mean$ calculates the mean of these ranks. A higher $Hits@k$ and a lower $Mean$ indicate better performance. It is a phenomenon worth noting that the optimal $Hits@k$ and $Mean$ usually do not come at the same epoch in all the three approaches. For fair comparison, we did not fix the number of epochs but used early stopping to avoid overtraining. The training process is stopped as long as the change ratio of $Mean$ is less than $0.0005$. Besides, the training of AE on each dataset takes 100 epochs.

\subsection{Experimental Results}
\label{subsect:result}

\subsubsection{Results on DBP15K}

We used a certain proportion of the gold standards as seed alignment while left the remaining as testing data, i.e., the latent aligned entities to discover. We tested the proportion from 10\% to 50\% with step 10\%, and Table~\ref{tab:dbp15k} lists the results using 30\% of the gold standards. The variation of $Hits@k$ with different proportions will be shown shortly. For relationships and attributes, we simply extracted the property pairs with exactly the same labels, which only account for a small portion of the seed alignment.

Table~\ref{tab:dbp15k} indicates that JAPE largely outperformed JE and MTransE, since it captures both structure and attribute information of KBs. For JE, it employs TransE as its basic model, which is not suitable to be directly applied to entity alignment as discussed in Section~\ref{subsect:discuss}. Besides, JE does not give a mandatory constraint on the length of vectors. Instead, it only minimizes $\|\mathbf{v}\|_2^2-1$ to restrain vector length and brings adverse effect. For MTransE, it models the structures of KBs in different vector spaces, and information loss happens when learning the translation between vector spaces.

\begin{table}[!tbp]
	\centering
	\caption{Result comparison and ablation study}
	\label{tab:dbp15k}  
	{\scriptsize  
		\subfloat{%
			\begin{tabular}{|l|l|cccc|cccc|} 
				\hline \multicolumn{2}{|l|}{\multirow{2}{*}{DBP15K$_\textrm{ZH-EN}$}} & \multicolumn{4}{|c|}{$\rm ZH\rightarrow EN$} & \multicolumn{4}{|c|}{$\rm EN\rightarrow ZH$} \\
				\cline{3-10}\multicolumn{2}{|c|}{} & $Hits@1$ & $Hits@10$ & $Hits@50$ & $Mean$ & $Hits@1$ & $Hits@10$ & $Hits@50$ & $Mean$ \\
				\hline \multicolumn{2}{|l|}{JE} & 21.27 & 42.77 & 56.74 & 766 & 19.52 & 39.36 & 53.25 & 841 \\
				\hline \multicolumn{2}{|l|}{MTransE} & 30.83 & 61.41 & 79.12 & 154 & 24.78 & 52.42 & 70.45 & 208 \\
				\hline \multirow{3}{*}{JAPE} & SE w/o neg. & 38.34 & 68.86 & 84.07 & 103 & 31.66 & 59.37 & 76.33 & 147 \\
				& SE          & 39.78 & 72.35 & 87.12 & 84  & 32.29 & 62.79 & 80.55 & 109 \\
				& $\rm SE+AE$ &\bf 41.18 &\bf 74.46 &\bf 88.90 &\bf 64 &\bf 40.15 &\bf 71.05 &\bf 86.18 &\bf 73 \\
				\hline
		\end{tabular}}
		
		\subfloat{%
			\begin{tabular}{|l|l|cccc|cccc|} 
				\hline \multicolumn{2}{|l|}{\multirow{2}{*}{DBP15K$_\textrm{JA-EN}$}} & \multicolumn{4}{|c|}{$\rm JA\rightarrow EN$} & \multicolumn{4}{|c|}{$\rm EN\rightarrow JA$} \\
				\cline{3-10}\multicolumn{2}{|c|}{} & $Hits@1$ & $Hits@10$ & $Hits@50$ & $Mean$ & $Hits@1$ & $Hits@10$ & $Hits@50$ & $Mean$ \\
				\hline \multicolumn{2}{|l|}{JE} & 18.92 & 39.97 & 54.24 & 832 & 17.80 & 38.44 & 52.48 & 864 \\
				\hline \multicolumn{2}{|l|}{MTransE} & 27.86 & 57.45 & 75.94 & 159 & 23.72 & 49.92 & 67.93 & 220 \\
				\hline \multirow{3}{*}{JAPE} & SE w/o neg. & 33.10 & 63.90 & 80.80 & 114 & 29.71 & 56.28 & 73.84 & 156 \\
				& SE          & 34.27 & 66.39 & 83.61 & 104 & 31.40 & 60.80 & 78.51 & 127 \\
				& $\rm SE+AE$ &\bf 36.25 &\bf 68.50 &\bf 85.35 &\bf 99 &\bf 38.37 &\bf 67.27 &\bf 82.65 &\bf 113 \\
				\hline
		\end{tabular}}
		
		\subfloat{%
			\begin{tabular}{|l|l|cccc|cccc|} 
				\hline \multicolumn{2}{|l|}{\multirow{2}{*}{DBP15K$_\textrm{FR-EN}$}} & \multicolumn{4}{|c|}{$\rm FR\rightarrow EN$} & \multicolumn{4}{|c|}{$\rm EN\rightarrow FR$} \\
				\cline{3-10}\multicolumn{2}{|c|}{} & $Hits@1$ & $Hits@10$ & $Hits@50$ & $Mean$ & $Hits@1$ & $Hits@10$ & $Hits@50$ & $Mean$ \\
				\hline \multicolumn{2}{|l|}{JE} & 15.38 & 38.84 & 56.50 & 574 & 14.61 & 37.25 & 54.01 & 628 \\
				\hline \multicolumn{2}{|l|}{MTransE} & 24.41 & 55.55 & 74.41 & 139 & 21.26 & 50.60 & 69.93 & 156 \\
				\hline \multirow{3}{*}{JAPE} & SE w/o neg.  & 29.55 & 62.18 & 79.36 & 123 & 25.40 & 56.55 & 74.96 & 133 \\
				& SE           & 29.63 & 64.55 & 81.90 & 95 & 26.55 & 60.30 & 78.71 & 107 \\
				& $\rm SE+AE$  &\bf 32.39 &\bf 66.68 &\bf 83.19 &\bf 92 &\bf 32.97 &\bf 65.91 &\bf 82.38 &\bf 97 \\
				\hline
		\end{tabular}}
		
		%\subfloat{%
		%\begin{tabular}{|ll|cccc|cccc|} 
		% \hline \multicolumn{2}{|c|}{\multirow{2}{*}{}} & \multicolumn{4}{|c|}{$\rm DE\rightarrow EN$} & \multicolumn{4}{|c|}{$\rm EN\rightarrow DE$} \\
		% \cline{3-10}\multicolumn{2}{|c|}{} &\ $Hits@1$\ &\ $Hits@10$\ &\ $Hits@50$\ &\ $Mean$\ &\ $Hits@1$\ &\ $Hits@10$\ &\ $Hits@50$\ &\ $Mean$\ \\
		% \hline JE      & & 17.14 & 36.37 & 49.59 & 1,086 & 15.62 & 33.86 & 47.87 & 1,121 \\
		% \hline MTransE & & 25.04 & 52.66 & 69.21 & 308 & 20.85 & 47.57 & 66.21 & 284 \\
		% \hline \multirow{3}{*}{JAPE} &\ SE w/o neg.\ & 29.54 & 59.05 & 75.45 & 253 & 25.11 & 53.63 & 71.67 & 248 \\
		%                              &\ SE           & 30.87 & 64.94 & 80.81 & 127 & 27.14 & 59.80 & 78.50 & 152 \\
		%        &\ $\rm SE+AE$ &\bf 37.03 &\bf 68.90 &\bf 83.83 &\bf 118 &\bf 38.37 &\bf 69.04 &\bf 84.22 &\bf 121 \\
		% \hline
		%\end{tabular}}
	}
\end{table}

Additionally, we divided JAPE into three variants for ablation study, and the results are shown in Table~\ref{tab:dbp15k} as well. We found that involving negative triples in structure embedding reduces the random distribution of entities, and involving attribute embedding as constraint further refines the distribution of entities. The two improvements demonstrate that systematic distribution of entities makes for the cross-lingual entity alignment task.

It is worth noting that the alignment direction (e.g. $\rm ZH\rightarrow EN$ vs. $\rm EN\rightarrow ZH$) also causes performance difference. As shown in Table~\ref{tab:stat}, the relationship triples in a non-English KB are much sparser than those in an English KB, so that the approaches based on the relationship triples cannot learn good representations to model the structures of non-English KBs, as restraints for entities are relatively insufficient. When performing alignment from an English KB to a non-English KB, we search for the nearest non-English entity as the aligned one to an English entity, the sparsity of the non-English KB leads to the disorganized distribution of its entities, which brings negative effects on the task. However, it is comforting to see that the performance difference becomes narrower when involving attribute embedding, because the attribute triples provide additional information to embed entities, especially for sparse KBs.

\begin{figure}[!tbp]
\centering
\includegraphics[width=\columnwidth]{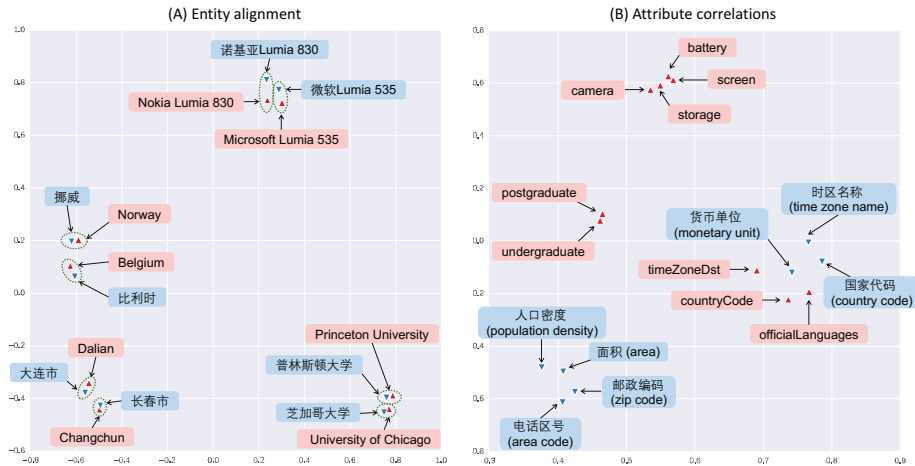}
\caption{Visualization of results on DBP15K$_\textrm{ZH-EN}$}
\label{fig:vis}
\end{figure}

Fig.~\ref{fig:vis} provides the visualization of sample results for entity alignment and attribute correlations. We projected the embeddings of aligned entity pairs and involved attribute embeddings to two dimensions using PCA. The left part indicates that universities, countries, cities and cellphones were divided widely while aligned entities from Chinese to English were laid closely, which met our expectation of JAPE. The right part shows our attribute embedding clustered three groups of monolingual attributes (about cellphones, cities and universities) and one group of cross-lingual ones (about countries).

\subsubsection{Sensitivity to Proportion of Seed Alignment}

\begin{figure}[!tbp]
	\centering
	\includegraphics[width=\columnwidth]{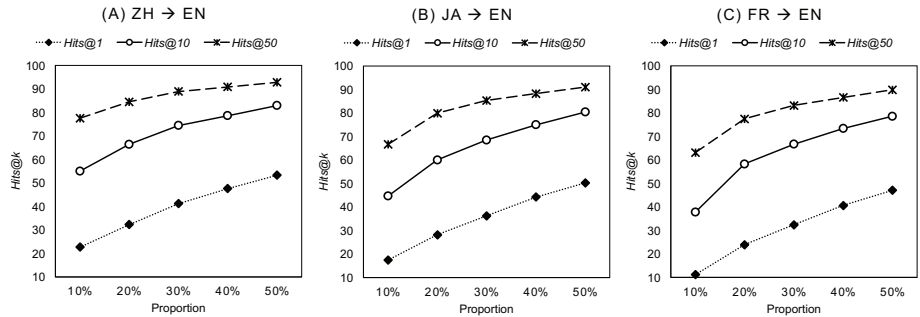}
	\caption{$Hits@k$ w.r.t. proportion of seed alignment}
	\label{fig:proportion}
\end{figure}

Fig.~\ref{fig:proportion} illustrates the change of $Hits@k$ with varied proportion of seed alignment. In accordance with our expectation, the results on all the datasets become better with the increase of the proportion, because more seed alignment can provide more information to overlay the two KBs. It can be seen that, when using half of the gold standards as seed alignment, JAPE performed encouragingly, e.g. $Hits@1$ and $Hits@10$ on DBP15K$_\textrm{ZH-EN}$ are 53.27\% and 82.91\%, respectively. Moreover, even with a very small proportion of seed alignment like $10\%$, JAPE still achieved promising results, e.g. $Hits@10$ on DBP15K$_\textrm{ZH-EN}$ reaches 55.04\% and on DBP15K$_\textrm{JA-EN}$ reaches 44.69\%. Therefore, it is feasible to deploy JAPE to various entity alignment tasks, even with limited seed alignment.

%\subsubsection{Case Study}

\subsubsection{Combination with Machine Translation} Since machine translation is often used in cross-lingual ontology matching \cite{MultilingualSW,CrosslingualOM}, we designed a machine translation based approach that employs Google Translate to translate the labels of entities in one KB and computes similarities between the translations and the labels of entities in the other KB. For similarity measurement, we chose Levenshtein distance because of its popularity in ontology matching \cite{StringSim}. 

We chose DBP15K$_\textrm{ZH-EN}$ and DBP15K$_\textrm{JA-EN}$, which have big barriers in linguistics. As depicted in Table~\ref{tab:comb}, machine translation achieves satisfying results, especially for $Hits@1$, and we think that it is due to the high accuracy of Google Translate. However, the gap between machine translation and JAPE becomes smaller for $Hits@10$ and $Hits@50$. The reason is as follows. When Google misunderstands the meaning of labels (e.g. polysemy), the top-ranked entities are all very likely to be wrong. On the contrary, JAPE relies on the structure information of KBs, so the correct entities often appear slightly behind. Besides, we found that translating from Chinese (or Japanese) to English is more accurate than the reverse direction.

To further investigate the possibility of combination, for each latent aligned entities, we considered the lower rank of the two results as the combined rank. It is surprising to find that the combined results are significantly better, which reveals the mutual complementarity between JAPE and machine translation. We believe that, when aligning entities between cross-lingual KBs where the quality of machine translation is difficult to guarantee, or many entities lack meaningful labels, JAPE can be a practical alternative.

\begin{table}[!tbp]
 \centering
 \caption{Combination of machine translation and JAPE}
 \label{tab:comb}  
 {\scriptsize  
  \subfloat{%
  \begin{tabular}{|l|cccc|cccc|} 
   \hline \multirow{2}{*}{DBP15K$_\textrm{ZH-EN}$} & \multicolumn{4}{|c|}{$\rm ZH\rightarrow EN$} & \multicolumn{4}{|c|}{$\rm EN\rightarrow ZH$} \\
   \cline{2-9} & $Hits@1$ & $Hits@10$ & $Hits@50$ & $Mean$ & $Hits@1$ & $Hits@10$ & $Hits@50$ & $Mean$ \\
   \hline Machine translation & 55.76 & 67.61 & 74.30 & 820 & 40.38 & 54.27 & 62.27 & 1,551 \\
   \hline JAPE                & 41.18 & 74.46 & 88.90 & 64 & 40.15 & 71.05 & 86.18 & 73 \\
   \hline Combination         &\bf 73.09 &\bf 90.43 &\bf 96.61 &\bf 11 &\bf 62.70 &\bf 85.21 &\bf 94.25 &\bf 26 \\
   \hline
  \end{tabular}}
   
  \subfloat{%
  \begin{tabular}{|l|cccc|cccc|} 
   \hline \multirow{2}{*}{DBP15K$_\textrm{JA-EN}$} & \multicolumn{4}{|c|}{$\rm JA\rightarrow EN$} & \multicolumn{4}{|c|}{$\rm EN\rightarrow JA$} \\
   \cline{2-9} & $Hits@1$ & $Hits@10$ & $Hits@50$ & $Mean$ & $Hits@1$ & $Hits@10$ & $Hits@50$ & $Mean$ \\
   \hline Machine translation & 74.64 & 84.57 & 89.13 & 333 & 61.98 & 72.07 & 77.22 & 1,095 \\
   \hline JAPE                & 36.25 & 68.50 & 85.35 & 99 & 38.37 & 67.27 & 82.65 & 113\\
   \hline Combination         &\bf 82.84 &\bf 94.65 &\bf 98.31 &\bf 9 &\bf 75.94 &\bf 90.70 &\bf 96.04 &\bf 25 \\
   \hline
  \end{tabular}}
 }
\end{table}

\subsubsection{Results at Larger Scale}

To test the scalability of JAPE, we built three larger datasets by choosing 100 thousand ILLs between English and Chinese, Japanese and French in the same way as DBP15K. The threshold of relationship triples to select ILLs was set to $2$. Each dataset contains several hundred thousand entities and several million triples. We set $d=100, \beta=0.1$ and keep other parameters the same as DBP15K. For JE, the training takes 2000 epochs as reported in its paper. The results on DBP100K are listed in Table~\ref{tab:dbp100k}. Due to lack of space, only $Hits@10$ is reported. We found that similar results and conclusions stand for DBP100K compared with DBP15K, which indicate the scalability and stability of JAPE. 

Furthermore, the performance of all the methods decreases to some extent on DBP100K. We think that the reasons are twofold: (i) DBP100K contains quite a few ``sparse" entities involved in a very limited number of triples, which affect embedding the structure information of KBs; and (ii) as the number of latent aligned entities in DBP100K are several times larger than DBP15K, the TransE-based models suffer from the increased occurrence of multi-mapping relations as explained in \cite{TransH}. Nevertheless, JAPE still outperformed JE and MTransE.

\begin{table}[!tbp]
 \centering
 \caption{$Hits@10$ comparison on DBP100K}
 \label{tab:dbp100k}  
 {\scriptsize  
  \subfloat{%
  \begin{tabular}{|l|cc|} 
   \hline DBP100K & $\rm ZH\rightarrow EN$ & $\rm EN\rightarrow ZH$ \\
   \hline JE      & 16.95 & 16.63 \\
   \hline MTransE & 34.31 & 29.18 \\
   \hline JAPE    &\bf 41.75 &\bf 40.13 \\
   \hline
  \end{tabular}}
  \subfloat{%
  \begin{tabular}{|cc|} 
   \hline $\rm JA\rightarrow EN$ & $\rm EN\rightarrow JA$ \\
   \hline 21.17 & 20.98 \\
   \hline 33.93 & 27.22 \\
   \hline \bf 42.00 &\bf 39.30 \\
   \hline
  \end{tabular}}
  \subfloat{%
  \begin{tabular}{|cc|} 
   \hline $\rm FR\rightarrow EN$ & $\rm EN\rightarrow FR$ \\
   \hline 22.98 & 22.63 \\
   \hline 44.84 & 39.19 \\
   \hline \bf 53.64 &\bf 50.51 \\
   \hline
  \end{tabular}}
 }
\end{table}

\section{Conclusion and Future Work}
\label{sect:concl}

In this paper, we introduced a joint attribute-preserving embedding model for cross-lingual entity alignment. We proposed structure embedding and attribute embedding to represent the relationship structures and attribute correlations of KBs and learn approximate embeddings for latent aligned entities. Our experiments on real-world datasets demonstrated that our approach achieved superior results than two state-of-the-art embedding approaches and could be complemented with conventional methods based on machine translation. 

In future work, we look forward to improving our approach in several aspects. First, the structure embedding suffered from multi-mapping relations, thus we plan to extend it with cross-lingual hyperplane projection. Second, our attribute embedding discarded attribute values due to their diversity and cross-linguality, which we want to use cross-lingual word embedding techniques to incorporate. Third, we would like to evaluate our approach on more heterogeneous KBs developed by different parties, such as between DBpedia and Wikidata.\\

\noindent\textbf{Acknowledgements.} This work is supported by the National Natural Science Foundation of China (Nos. 61370019, 61572247 and 61321491).
%--------------------

\bibliographystyle{splncs03}
\bibliography{reference2}

\end{document}